\documentclass[letterpaper, 10 pt, journal, twoside]{IEEEtran}

\makeatletter
\let\NAT@parse\undefined
\makeatother

\IEEEoverridecommandlockouts 
\usepackage{multicol}
\usepackage{multirow}

\usepackage[utf8]{inputenc} %
\usepackage[T1]{fontenc}    %
\usepackage{hyperref}       %
\usepackage{url}            %
\usepackage{booktabs}       %
\usepackage{amsfonts}       %
\usepackage{nicefrac}       %
\usepackage{xcolor}         %
\usepackage{xspace}
\usepackage{soul}

\usepackage{wrapfig}

\usepackage[numbers,sort&compress]{natbib}
\usepackage{amssymb}
\usepackage{wrapfig}
\usepackage{subfigure}
\usepackage{caption}
\usepackage{graphicx}
\usepackage{amsmath}

\usepackage{tikz-cd}

\usepackage{colortbl}
\usepackage{algorithm}
\usepackage[noend]{algorithmic}
\usepackage{bbding}
\usepackage{fancyvrb}
\usepackage{listings}
\definecolor{codegreen}{rgb}{0,0.5,0}
\definecolor{codered}{rgb}{0.7,0.1,0.1}
\definecolor{codegray}{rgb}{0.5,0.5,0.5}
\definecolor{codepurple}{rgb}{0.58,0,0.82}
\definecolor{backcolour}{rgb}{0.95,0.95,0.95}
\lstdefinestyle{python}{
    language=Python,
    basicstyle=\ttfamily\scriptsize,
    backgroundcolor=\color{backcolour},   
    commentstyle=\color{codegreen}\ttfamily\slshape,
    keywordstyle=\color{codered}\bfseries,
    numberstyle=\tiny\color{codegray},
    stringstyle=\color{codepurple},    
    xleftmargin=1em,
    framexleftmargin=0.3em,
    breakatwhitespace=false,         
    breaklines=true,                 
    captionpos=b,  %
    keepspaces=true,                 
    numbers=left,                    
    numbersep=4pt,                  
    showspaces=false,                
    showstringspaces=false,
    showtabs=true,                  
    tabsize=1,
    fancyvrb=true
}
\usepackage[disable,textsize=tiny]{todonotes}

\newcounter{todocounter}

\newcommand{\todoil}[1]{\todo[inline, size=\small, color=orange!50]{#1}}
\newcommand{\todozlf}[2][]{\todo[color=yellow!80!black, #1]{LF: #2}}
\newcommand{\todoilzlf}[2][]{\todo[inline, size=\small, color=yellow!80!black, #1]{LF: #2}}
\newcommand{\todoref}[1]{\todo[color=blue!30]{#1}}  %
\newcommand{\todonote}[1]{\todo[color=blue!50!white]{#1}}

\usepackage{amsmath,amsfonts,bm}

\def\eqref#1{equation~\ref{#1}}

\def\1{\bm{1}}

\def\va{{\bm{a}}}

\def\vh{{\bm{h}}}

\def\vm{{\bm{m}}}

\def\vs{{\bm{s}}}

\def\vw{{\bm{w}}}
\def\vx{{\bm{x}}}

\def\mK{{\bm{K}}}

\DeclareMathAlphabet{\mathsfit}{\encodingdefault}{\sfdefault}{m}{sl}
\SetMathAlphabet{\mathsfit}{bold}{\encodingdefault}{\sfdefault}{bx}{n}

\def\gA{{\mathcal{A}}}

\def\gE{{\mathcal{E}}}

\def\gG{{\mathcal{G}}}

\def\gS{{\mathcal{S}}}
\def\gT{{\mathcal{T}}}

\def\gV{{\mathcal{V}}}

\def\sR{{\mathbb{R}}}

\def\sZ{{\mathbb{Z}}}

\newcommand{\editcorl}[1]{{\color{blue} #1}}

\newcommand{\ethree}{{\mathrm{E}(3)}}

\newcommand{\etwo}{{\mathrm{E}(2)}}

\newcommand{\sotwo}{{\mathrm{SO}(2)}}
\newcommand{\otwo}{{\mathrm{O}(2)}}

\DeclareMathOperator{\Hom}{Hom}

\DeclareMathOperator{\Res}{Res}

\def\eg{\textit{e.g.}\@\xspace}

\title{
$\mathrm{E}(2)$-Equivariant Graph Planning for Navigation
}

\begin{document}
\bstctlcite{BSTcontrol}

\author{Linfeng Zhao$^{1*}$, Hongyu Li$^{1,2*}$, \textit{Student Member, IEEE,} Ta\c{s}k{\i}n Pad{\i}r$^{1}$, \textit{Senior Member, IEEE,}\\
Huaizu Jiang$^{1 \dag}$, \textit{Member, IEEE}, and Lawson L.S. Wong$^{1 \dag}$, \textit{Member, IEEE} %
\thanks{Manuscript received: September, 20, 2023; Revised December, 10, 2023; Accepted January, 19, 2024.}%
\thanks{This paper was recommended for publication by Editor Hanna Kurniawati upon evaluation of the Associate Editor and Reviewers' comments.
This work is supported by NSF Grants \#2107256 and \#2142519.} %
\thanks{$*$ equal contribution. $\dag$ equal advising.}
\thanks{$^{1}$ Linfeng Zhao, Ta\c{s}k{\i}n Pad{\i}r, Huaizu Jiang, and Lawson L.S. Wong are with Northeastern University, Boston, MA, USA, 02115.
        {\tt\small \{zhao.linf, t.padir, h.jiang, l.wong\}@northeastern.edu}.}%
\thanks{$^{2}$ Hongyu Li is with Brown University, Providence, RI, USA, 02912  {\tt\small hongyu@brown.edu}. This work was partially done when Hongyu Li was at Northeastern University.}
\thanks{Digital Object Identifier (DOI): see top of this page.}
}

\markboth{IEEE Robotics and Automation Letters. Preprint Version. Accepted January, 2024}
{Zhao \MakeLowercase{\textit{et al.}}: $\mathrm{E}(2)$-Equivariant Graph Planning for Navigation} 

\maketitle

\begin{abstract}
    Learning for robot navigation presents a critical and challenging task. The scarcity and costliness of real-world datasets necessitate efficient learning approaches. 
    In this letter, we exploit Euclidean symmetry in planning for 2D navigation, which originates from Euclidean transformations between reference frames and enables parameter sharing.
    To address the challenges of unstructured environments,
    we formulate the navigation problem as planning on a geometric graph and develop an equivariant message passing network to perform value iteration.
    Furthermore, to handle multi-camera input, we propose a learnable equivariant layer to lift features to a desired space.
    We conduct comprehensive evaluations across five diverse tasks encompassing structured and unstructured environments, along with maps of known and unknown, given point goals or semantic goals.
    Our experiments confirm the substantial benefits on training efficiency, stability, and generalization.
    More details can be found at the project website: \url{https://lhy.xyz/e2-planning/}.
\end{abstract}


\section{Introduction}

\IEEEPARstart{N}{avigation} is a fundamental capability of mobile robots. Traditional navigation approaches, such as A* \cite{hart_formal_1968}, focus on finding shortest-distance collision-free paths to a provided goal location in a pre-built occupancy map or known costmap.
Recently, learning-based approaches to robot navigation have been proposed \cite{chaplot_object_2020,chang_semantic_2020, liang_sscnav_2021,akmandor_deep_2022, brohan_rt-1_2022}, which are particularly useful when the costs or goals are not explicitly provided and need to be learned from data. For example, in visual navigation, the cost to navigate between locations may depend on high-dimensional visual features, and the goal may likewise need to be visually identified (\eg, ``find a mug''). As another example, in imitation learning, users may provide information about their preferred navigation policy implicitly via demonstrations, and the costs or optical actions need to be learned using features from the robot's state-action space.

\begin{figure*}[!t]
    \vspace*{0.15cm} %
    \begin{center}
        \includegraphics[width=.75\textwidth]{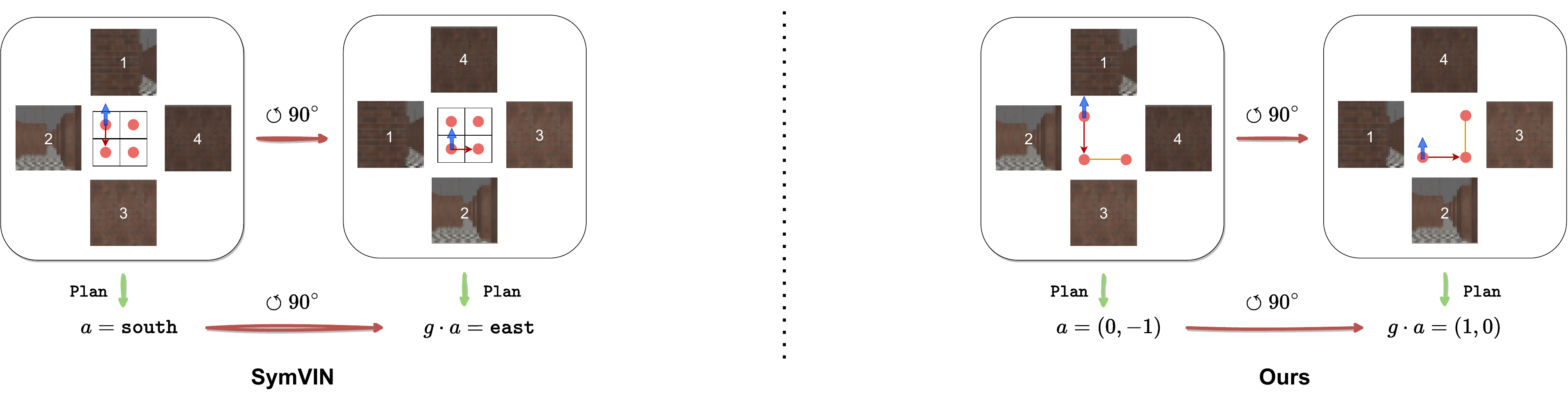}
    \end{center}
  \caption{
  \small
  \textbf{Illustration of rotation equivariance.}
  We provide a side-by-side comparison with SymVIN \cite{zhao_integrating_2022}.
  We use the {\color{blue} blue} arrow to show the orientation of the robot.
  Rotating the robot $\circlearrowright 90^{\circ}$ is equivalent to rotating the world frame $\circlearrowleft 90^{\circ}$.
  When camera views are cyclically permuted, action output (red arrow) is transformed by a rotation matrix.
  The state space of SymVIN (left) is confined to the grid, and it only produces discrete actions. Our approach acts on continuous 2D space and produces $\sR^2$ actions.
  }
    \label{fig:fig1-commutative}
\end{figure*}

While the aforementioned learning-based approaches exhibit remarkable capability in handling high-dimensional observations, they typically require a considerable amount of data and intensive training \cite{chaplot_object_2020,chang_semantic_2020}. Furthermore, these methods lack guarantees regarding generalization capabilities. In this work, we investigate the potential benefits of Euclidean symmetry in navigation tasks. It stems from Euclidean transformations among reference frames, enabling parameter sharing, enhancing efficiency, and improving generalizability.
The utilization of symmetry in navigation within the \emph{grid world domain} is explored in the earlier study by \citet{zhao_integrating_2022} (left of Fig.~\ref{fig:fig1-commutative}). They introduce the equivariant version of the value iteration network (VIN) \cite{tamar_value_2016} under \textit{discrete translations, rotations, and reflections}, along with a differentiable navigation planner. Their work showcases notable improvements compared to baseline approaches \cite{tamar_value_2016, lee_gated_2018_hongyu}. However, they only focused on navigation in discrete 2D grids, which limits its applicability to robot navigation.

In our work, we introduce an equivariant learning-based navigation approach that operates on graphs in continuous space and considers symmetry with respect to an infinitely larger continuous group -- the Euclidean group $\etwo$ (right of Fig.~\ref{fig:fig1-commutative}). Specifically, we use \textit{geometric graphs} (or spatial graphs) \citep{bronstein_geometric_2021}, where nodes in our graph correspond to states (and their features) arbitrarily located in 2D space. This eliminates the confinement to a grid, enabling the environment to remain non-discretized and permitting variable resolution. This also helps when the robot's motion deviates from grid-like patterns. Moreover, our approach accounts for continuous rotational symmetry, enhancing learning efficiency compared to discrete symmetry like Dihedral group $D_4$.

However, to exploit Euclidean symmetry in graph-based navigation, we need to solve two major challenges. First, previous work on 2D grids exploited the grid nature of their problem and used standard 2D symmetric convolution, which is no longer applicable in our case. Instead, we derive a new $\etwo$-equivariant message-passing version of VIN and validate that it satisfies our notion of symmetry.
Second, to capture symmetry in visual inputs/features, previous work relied on a very specific setup.
As illustrated in Fig. \ref{fig:fig1-commutative},
the agent was assumed to have four cameras, each situated $90^\circ$ apart, exactly matching the $D_4$ symmetry being considered, such that a group transformation (rotation) can be implemented as a permutation to the four images. Extending this approach directly to $\etwo$ would technically require an infinite number of cameras (or at least an infinite-resolution panoramic camera).
We lift this restriction by introducing a learnable equivariant layer that can take images from a camera array conforming to a subgroup of $\etwo$ (such as $D_8$) and lift their features to become $\etwo$-equivariant.

We empirically demonstrate the effectiveness of our approach on various navigation environments, including 2D grid, 2D geometric graphs, and Miniworld visual navigation \cite{MinigridMiniworld23} on both grid and graph. 
Moreover, in demonstrating its potential suitability for semantic goals and real-world environments, we provide a proof-of-concept experiment on semantic navigation tasks in the Habitat simulator \cite{savva_habitat_2019_hongyu}. Among these studies, we observe a consistent improvement in learning efficiency and stability.
Overall, our study provides insight into the application of equivariance in navigation and the challenges.
Our contributions are three-fold: 
\begin{itemize}
	\item We study the equivariance properties of 2D navigation and identify the two challenges.
        \item To address the challenges, we (1) derive the geometric message passing (MP) version of value iteration on geometric graphs and (2) propose using a learnable equivariant layer that converts multi-camera images to desired feature space, respectively.
	\item We demonstrate the empirical performance of navigation on Grid World (2D grid), Graph World (2D geometric graphs), and Miniworld visual navigation on both grid and graph. We provide proof-of-concept results on semantic navigation in Habitat simulator.
\end{itemize}

\section{Related Works}

\textbf{Geometric deep learning.}
Our exploration of Euclidean symmetry utilizes tools from geometric deep learning \citep{bronstein_geometric_2021,cohen_group_2016_hongyu,satorras_en_2021_hongyu,brandstetter_geometric_2021_hongyu,weiler_general_2021_hongyu}.
Geometric deep learning and equivariant networks extend the study of classic 2D translation-equivariant convolution neural networks into more symmetry groups and spaces \citep{bronstein_geometric_2021,cohen_group_2016_hongyu}.
\citet{cohen_group_2016_hongyu} propose group convolution network (G-CNN), a pioneer work that studies rotation symmetry, followed by an extension to steerable convolution, Steerable CNN \citep{cohen_steerable_2016_hongyu}.
It has also been extended to the 3D case \citep{weiler_3d_2018} and supported by a library in $\etwo$ \citep{weiler_general_2021_hongyu}.
For graphs, equivariant message passing uses equivariant multilayer perceptrons (MLPs) to propagate geometric quantities between nodes to preserve the symmetry \citep{satorras_en_2021_hongyu,brandstetter_geometric_2021_hongyu}.
Different from $\ethree$-equivariant message passing in \citep{brandstetter_geometric_2021_hongyu}, we work on $\etwo$ case.
Additionally, the relationship between geometric graphs and value iteration has been discussed in \citep{dudzik_graph_2022_hongyu}.
In practice, equivariant networks enable sharing parameters and reduce the number of parameters.

\textbf{Equivariance in reinforcement learning and planning.}
Our work draws upon previous research on symmetry in reinforcement learning (RL) and planning \citep{van2020mdp,wang_mathrmso2-equivariant_2021_hongyu,zhao_integrating_2022}.
Symmetry and equivariance have been studied in reinforcement learning and planning before and in the era of deep learning \citep{fox_extending_2002, pochter_exploiting_2011}.
Invariance of the optimal value function and equivariance of the optimal policy of a Markov Decision Process (MDP) with symmetry have been shown in \citet{zinkevich_symmetry_2001_hongyu}.
When using function approximation, equivariant policy networks and invariant value networks have been used to improve training efficiency in model-free RL \citep{van2020mdp,wang_mathrmso2-equivariant_2021_hongyu}, and equivariance also helps in transition model and model-based RL \citep{zhao2022toward,park2022learning,zhao_integrating_2022,zhao2023frame}.

\textbf{Learning to navigate.}
To achieve end-to-end navigation learning, several works investigate the differentiable planning algorithms~\cite{tamar_value_2016, yonetani_path_2021}.
In this letter, we aim to investigate a particular class of planning algorithms that rely on Value Iteration Network (VIN) \citep{tamar_value_2016} and its variants \citep{lee_gated_2018_hongyu,zhao_integrating_2022,zhao2023scaling,zhao2020nav}. 
The selection of VINs is motivated by the fact that value iteration is fully differentiable and inherently encompasses an equivariant convolution \citep{zhao_integrating_2022}.
\citet{gupta_cognitive_2017} adapt VIN to real-world applications with simultaneous mapping and planning, and \citet{karkus_differentiable_2019_hongyu} propose DAN for end-to-end learning with structured representation. 
Prior to us, \citet{zhao_integrating_2022} improved VIN with symmetry.
However, these works operate on a structured 2D grid $\sZ^2$.
In this letter, we extend the planning to the 2D plane, enabling navigation in more realistic unstructured environments.

\section{Background and Problem Formulation: Navigation as Geometric Graphs}

In this section, we define the navigation problem under study and explore its symmetry aspects.
Our formulation is a straightforward generalization of the global planning on occupancy grid \citep{tamar_value_2016,zhao_integrating_2022}, with extensions including representing the navigation task through a \textit{geometric graph} \citep{bronstein_geometric_2021,brandstetter_geometric_2021_hongyu}.
Our objective is to train a planner that generates action $\va_t$ at state $\vs_t$, guiding the agent to reach a target $\vw$ on the graph: $\va_t = \texttt{policy}_\theta(\vs_t, \vw)$.
The target can be a spatial location (point goal) or semantic goal.
We base on the differentiable planner -- VIN \citep{tamar_value_2016}, allowing to consume input in high-dimensional features, \eg, image or even text embedding.
As background, we first explain the problem definition, alongside the geometric structure and symmetry in the navigation graph. 
Then, we introduce the extension of equivariance in value iteration. Lastly, we delve into the incorporation of equivariance within the value iteration framework on the geometric graph.

\textbf{Definition.}
We approach navigation as a 2D continuous path planning problem, building upon the 2D discrete grid version introduced in \citep{tamar_value_2016,zhao_integrating_2022}, while extending it to the utilization of the \textit{geometric graph} $\gG = \langle \gV, \gE \rangle$ in 2D Euclidean space $\sR^2$.
In navigation tasks, the agent observes a state $\vs_t \in \gS$ at each step, and the action is to move on the 2D plane $\va_t \in \gA = \sR^2$. State $\vs_t$ can be a 2D position in $\sR^2$ or egocentric panoramic images in $\sR^{K\times H \times W}$ (where $K$ denotes the number of images of resolution $H \times W$)\footnote{We omit image RGB channel for notation simplicity.}. 
To convert the task into a geometric graph, each node $v_i \in \gV$ corresponds to a state $\vs \in \gS$ and is associated with a \textit{node feature} $\vh_i$ (such as images) and has a position $\vx_i \in \sR^2$.
It is also possible to use \textit{edge features}\footnote{
Similarly, each edge $e_{ij} \in \gE$ corresponds to a state-action transition $(\vs,\va) \in \gS \times \gA$ and has an \textit{edge feature} $ \in \sR^{c_e}$ (such as distance or movement cost).
}.
In this letter, we focus on addressing the global planning problem: given a navigation task (state $\vs$ and target $\vw$) as a feature field/map $M$, we output 
\todonote{(fixed) LSW: May not be clear what an action field is; agree with inline comment. Using both $A$ and $\mathcal{A}$ is confusing.} action field $\Pi = \texttt{Plan}_\theta(M)$.

\textbf{Assumptions.}
The navigation challenge under consideration pertains to high-level global planning. In this context, we abstract the perception aspect (\eg, the method of acquisition) and the control aspect (assuming the feasibility of 2D relative movement output).
Even if the execution of action does not arrive at another graph node, we may use action from the closest states or interpolate surrounding states.
We assume a relatively accurate localization is provided.

\textbf{Geometric Structure.}
The navigation problem can be defined as a MDP, and an inherent geometric structure emerges:
it can be conceptualized as a geometric graph (defined above) situated within \textit{a 2D Euclidean space}.
Specifically, this graph can be transformed through 2D Euclidean isometric symmetries, without impacting the optimal solution of the MDP \cite{van2020mdp,zhao_integrating_2022}.
The set of all such transformations in 2D is called \textit{Euclidean group} $\mathrm{E}(2)$, which can be uniquely decomposed into translation part $\sR^2$ and rotation/reflection part $\mathrm{O}(2)$, denoted as semi-direct product $\rtimes$: $\mathrm{E}(2) = \sR^2 \rtimes \mathrm{O}(2)$ \citep{cohen_group_2016_hongyu,weiler_general_2021_hongyu}.
We only require the node features $\vh$ (and edge features) are transformable by a subgroup $G \le \etwo$.
\todozlf{clean up}
Following the notation in \citep{weiler_general_2021_hongyu}, we denote the rotation/reflection symmetry part as compact symmetry group $G \leq \mathrm{GL}(2)$, because translation group is \textit{not compact} and many useful theorems do not hold.
\todo{what theorems? give some examples or remove?}
In our implementation, translation equivariance is achieved by using \textit{relative position}.
For any subgroups $G$ of rotation/reflection, its equivariance needs \textit{group convolution} \citep{cohen_group_2016_hongyu} or \textit{steerable convolution} \citep{cohen_steerable_2016_hongyu}.

\textbf{Value Iteration and Symmetry.}
When symmetry appears in an MDP, the value and policy functions are equivariant \citep{van2020mdp,zhao_integrating_2022}.
\todoref{check cite}
Abstractly, we can write value iteration (\texttt{VI}) as iteratively applying the Bellman operator $\gT: V_{t} \mapsto V_{t+1}$:
\begin{equation}
\begin{aligned}
\label{eqn: value-iteration}
    Q_{t}(\vs, \va) &:= R(\vs, \va) + \int_{\sR^2} d\vs' P(\vs' \mid \vs, \va) V(\vs'), \\
    \quad V_{t+1}(\vs) &= \max_{\va}Q_{t}(\vs, \va),
\end{aligned}
\end{equation}
where the input and output of the Bellman operator are both value function $V: \gS \to \sR$. Specifically, $\vs \in \gS, \va \in \gA, R(\vs, \va), P(\vs' \mid \vs, \va)$ represent the states, actions, rewards, and transitions, respectively.
In VIN, $\texttt{VI}(M) = \gT^k_M \left[ V_0 \right]$ is executed $k$ times, which takes an initial value $V_0$ and the map $M$ (with a goal) as input
\begin{equation}
\begin{aligned}
    \quad g \cdot \texttt{VI}(M) & \equiv g \cdot  \mathcal{T}^k_M [V_0] = \mathcal{T}^k_M [g \cdot V_0] \equiv \texttt{VI}(g \cdot M).
\end{aligned}
\end{equation}
\citet{zhao_integrating_2022} explore the equivariance for a 2D grid case.
We extend it to geometric graph: $\gT$ is performed on a graph, which is implemented using message passing.

\textbf{Symmetry Transformations.}
In this paragraph, we unify the concepts presented in the preceding two paragraphs to demonstrate the implementation of equivariance constraints, which establish equivalence between transformed and original input/output \citep{cohen_group_2016_hongyu,cohen_steerable_2016_hongyu,lang_wigner-eckart_2020_hongyu}.
\todo{explain more -> consider to move this to appendix; or extend for rebuttal; then move to appendix}
Under the group transformation $g$, a (left) \textit{regular} representation $L_g$ transforms a feature map with $c_\text{out}$-dimensional vector (vector field) $f: X \to \sR^{c_\text{out}}$ as~\citep{cohen_steerable_2016_hongyu,lang_wigner-eckart_2020_hongyu,bronstein_geometric_2021}:
\begin{equation}
	\left[L_g f\right](x)=\left[f \circ g^{-1}\right](x)= \rho_\text{out}(g) \cdot f\left(g^{-1} x\right),
\end{equation}
where $\rho_\text{out}$ is the $G$-representation associated with output $\sR^{c_\text{out}}$.
For example, for \textit{action} feature map $\Pi: \sR^2 \to \sR^2$ (i.e., every position $\vx  \in \sR^2$ is associated with a relative 2D movement), rotating the vector needs a $2 \times 2$ rotation matrix.

There are several useful functions in reinforcement learning (RL) and planning that can be written as graph features, e.g., node features as functions on $\gS$ and edge features as functions on $\gS  \times \gA$.
We use $\rho_\gS(g)$ to represent how the state is transformed under rotations and reflections $g \in G$, and similarly for action associated with representation $\rho_\gA(g)$.
\todozlf{explain more, potentially another paragraph; or appendix - add explanation of group representations; at least trivial,standard,regular}
Note that $M$ and $\Pi$ are \textit{vector} maps, requiring additional transformation for their respective fibers (vectors).
\todozlf{explain fiber}
When $\gA$ is continuous action, $\rho_\gA$ is rotation matrices.
For image-input case of $M: \gS \to \sR^{K\times H \times W}$, $\rho_{\text{camera}}(g)$ means cyclically permuting $K$ cameras: $\rho_{\text{camera}}(g) \cdot {M}(\vs_{t} ) =  {M}(\rho_\gS(g) \cdot \vs_{t})$.
It will be discussed in the next section.
We list the equivariance conditions of the key MDP functions here.

\vspace{-1.5em}
\begin{equation}
\begin{aligned}
R &: \gS \times \gA \to \sR :& {R}(\vs_{t}, \va_{t} ) =  {R}(\rho_\gS(g) \cdot \vs_{t}, \rho_\gA(g) \cdot \va_{t})  \\
Q &: \gS \times \gA \to \sR :& {Q}(\vs_{t}, \va_{t} ) =  {Q}(\rho_\gS(g) \cdot \vs_{t}, \rho_\gA(g) \cdot \va_{t})  \\
V &: \gS \to \sR :& {V}(\vs_{t} ) =  {V}(\rho_\gS(g) \cdot \vs_{t})  \\
\Pi &: \gS \to \gA :& \rho_\gA(g) \cdot \Pi (\vs_{t} ) = \Pi(\rho_\gS(g) \cdot \vs_{t}) 
\end{aligned}
\end{equation}

\section{Methodology: Equivariant Message Passing for Value Iteration}
\label{sec: methodology}

\begin{figure*}[t]
\vspace*{0.15cm}
\centering
\subfigure{
\includegraphics[width=.75\linewidth]{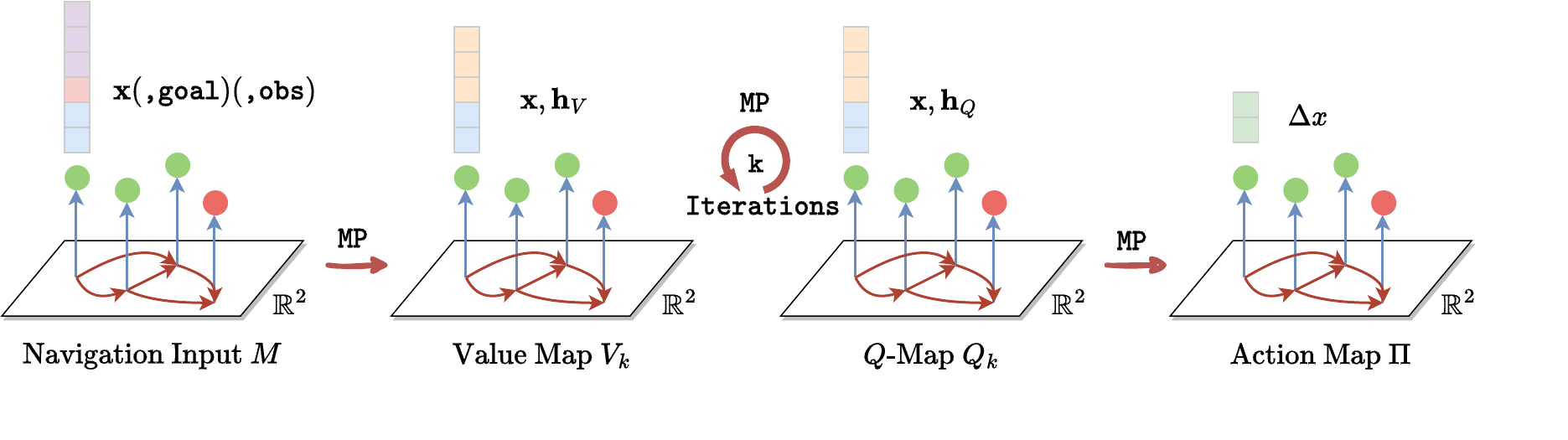}
}
\centering
\caption{
\small
\textbf{Overview of the message passing planner network (MP-VIN).} 
It takes the map $M$ as the input,
which contains the node position $\vx \in \sR^2$ and is optionally appended by the goal information (the {\color{red} red} node is goal node) \textit{or} observations depending on the navigation task.
Then, the output is applied value iteration for $k$ times. 
The state value map $h_V$ and Q-value map $h_Q$ are updated during value iterations.
The final output is an action map $\Pi$: for each node, it is a continuous relative movement $\Delta x \in \sR^2$.
}
\vspace{-15pt}
\label{fig:overview-mp}
\end{figure*}

Following the spirit of VIN, we build a geometric message passing network and extend it to learning value iteration on geometric graphs: $\Pi = \texttt{Plan}_\theta(M)$.
Given that the input feature map $M$ and the resulting action map $\Pi$ are both amenable to transformation within the same group, we enforce equivariance constraints throughout the MP network (shown in Fig.~\ref{fig:fig1-commutative}):
$ g \cdot \Pi = \texttt{Plan}_\theta(g \cdot M)$.

\subsection{Message Passing Value Iteration Networks (MP-VIN)}
The overview of the MP-VIN is shown in Fig.~\ref{fig:overview-mp}. For the input feature map $M$, each node contains a node position $\vx \in \sR^2$. Varied navigation tasks may lead to the augmentation of additional features, including goal features, observations, or a combination of both. For example, in the Grid World experiment (Sec.~\ref{sec: grid-world}), only the goal feature (a boolean value) is provided for each node. In the Miniworld experiment (Sec.~\ref{sec: mini-world}), both the goal feature and egocentric RGB observation are provided. In the semantic goal navigation experiment (Sec.~\ref{sec: habitat}), only the RGB observation is provided, rendering the goal implicit.
\todonote{(fixed) LSW: Do we ever use edge features? If not, maybe remove from formulation.}

Regarding the value iteration process, our MP-VIN is analogous to the original VIN formulation. However, what sets our approach apart is the improvement brought about by the inclusion of the geometric graph using an equivariant message passing layer (discussed in the next section). 
There are two advantages of using graph format: (1) cover the environment with irregular graphs to achieve variable resolution, and (2) output continuous actions.

\subsection{$\otwo$-Equivariant Message Passing Layer: Equivariant Value Iteration on Graph}

\textbf{Discretization to Graph.}
We could employ standard 2D convolution on regular grids for value iteration, as seen in VIN. However, irregular graphs render grids unsuitable.
In the prior study of \citet{niu_generalized_2017_hongyu}, an earlier iteration of graph convolution was employed.  However, it exhibited equivariance only with respect to $\sR^2$ translations, and it did not encompass considerations for rotation or reflection symmetries ($\otwo$).
Here, we derive from first principles using the original continuous form of value iteration.
\todo{add explanation: $\vh$ is a feature field, $\vh(\vx)$ is a value taken at a specific point}
\todo{to clean up notation; $\vh$ means feature field $\vh: \sR^2 \to \sR^c$, $\vh(\vx)$ means vector $\sR^c$}

The integral term in VI can be written as a mapping $\Phi$
\begin{equation}
    \vh'(\vx) = \Phi[h](\vx) = \int_{\sR^2} \mK (\vx, \vx') \vh (\vx'),
\end{equation}
where $\mK: \sR^2 \times \sR^2 \to \sR^{c_\text{out} \times c_\text{in}}$ is the kernal function \footnote{Note that the kernel $\mK$ here is different from the notation $K$ we use to represent the number of images.}. $\vh: \sR^2 \to \sR^{c_\text{in}}$ and $\vh': \sR^2 \to \sR^{c_\text{out}}$ are input and output feature map.
\todozlf{(close to fix) What is K? Any name? We should change a notation since $\sR^{K\times H \times W}$ - okay use $\mK$}
\todozlf{explain difference in 2-arg and 1-arg $\mK$}
If translation equivariance is desired, the kernel can be further simplified from two-argument to one-argument case, and the mapping is \textit{convolution} \citep{cohen_general_2020_hongyu,zhao_integrating_2022}.
\todozlf{convolution - translation equivariance - only depends on $x' - x$}
The continuous steerable convolution $\star$ is defined (via cross-correlation) by \citep{cohen_steerable_2016_hongyu,weiler_general_2021_hongyu,lang_wigner-eckart_2020_hongyu}:
\begin{equation}
\label{eqn: cont-steer-conv}
    \vh'(\vx)  = \left[ \mK \star \vh \right] (\vx) = \int_{\sR^2} \mK (\vx' - \vx) \vh (\vx'),
\end{equation}
where $\mK: \sR^2 \to \sR^{c_\text{out} \times c_\text{in}}$ is a (steerable) kernel.
\todozlf{add steerable constraint?}

If we sample nodes in $\sR^2$ and construct edges by transition $\gS \times \gA$, the continuous convolution on $\sR^2$ can be discretized, which is similar to strategy of PointConv \citep{brandstetter_geometric_2021_hongyu} %
\footnote{
\citet{brandstetter_geometric_2021_hongyu} discuss other strategy for 3D steerable messsage passing, which expands the feature maps to spherical harmonics.
Analogously, it is possible to expand the features to cyclic harmonics.
}.
We use \textit{nonlinear message passing} to replace linear convolution.
We use two MLPs for computing messages ($\texttt{propagate}_\theta$) and updating node features ($\texttt{update}_\theta$), and has form
\begin{equation}
\begin{aligned}
\vm_{ij} &= \texttt{propagate}_\theta \left( \vh_i, \vh_j, \vx_i, \vx_j\right), \\
\quad \vh'_i &= \texttt{update}_\theta \left(\vh_i, \sum_{j \in \mathcal{N}(i)} \vm_{ij}  \right).
\end{aligned}
\end{equation}

\textbf{Implementation of Equivariance.}
We implement $\etwo$-equivariant message passing on the graph that is equivariant under two parts: 

\textit{Translation $\sR^2$}.
In the plannar convolution on 2D grid, it is known to be equivariant to translation because it only relies on relative position between two cells as input and never takes absolute coordinates.
Analogously, we use relative position between nodes $\vx_i - \vx_j$ as input to the message passing function~\citep{brandstetter_geometric_2021_hongyu}:
\begin{equation}
    \vm_{ij} = \texttt{propagate}_\theta \left( \vh_i, \vh_j, \vx_i - \vx_j\right).
\end{equation}
It is a direct generalization of translation-equivariant 2D convolution that relies only on relative positions or local coordinates (shown in Eq.~\ref{eqn: cont-steer-conv}), allowing generalization to larger maps.

\textit{Rotation and Reflection $\otwo$}. 
We use steerable equivariant network to implement $\otwo$-equivariance 
\citep{cohen_steerable_2016_hongyu,cohen_group_2016_hongyu,weiler_general_2021_hongyu,brandstetter_geometric_2021_hongyu,cohen_general_2020_hongyu}. %
The $\otwo$ group is compact and thus its representations are decomposable into \textit{irreducible representations} \citep{weiler_general_2021_hongyu,lang_wigner-eckart_2020_hongyu}, thus convolutions  can be performed in Fourier domain and more efficient.
We use it to build equivariant MLPs of \texttt{propagate} and \texttt{update} (effectively $1 \times 1$ convolution). \todo{briefly introduce the architecture of these MLPs}
The kernel $\mK$ of $G$-steerable convolution needs to satisfy constraint \citep{weiler_general_2021_hongyu,lang_wigner-eckart_2020_hongyu}, where $G$ can be any (discrete) subgroup of $\otwo$:
\begin{equation}
\begin{aligned}
 	\mK(g x)=\rho_{\text {out }}(g) \circ \mK(x) \circ \rho_{\text {in }}(g)^{-1}
        \quad \forall g \in G, x \in \mathbb{R}^2,
\end{aligned}
\end{equation}
where $\rho_{\text {in}}$ and $\rho_{\text {out }}$ stand for representations of the layer's input and output, respectively.
This kernel constraint guarantees that the layer is $G$-equivariant: $\mK(g x) \rho_{\text {in }}(g)=\rho_{\text {out }}(g) \circ \mK(x)$.
We refer the readers to \citet{cohen_steerable_2016_hongyu,weiler_general_2021_hongyu} for more details.

\subsection{$C_K$-Equivariant Lifting Layer: Processing Camera Array}

In the previous section, we extend from discrete symmetry in SymVIN to continuous symmetry, such as continuous rotations $\sotwo$.
Injecting such equivariance into the \textit{entire} network requires us to know how to \textbf{continuously} rotate sensory input by $g \in SO(2)$. This can be naturally achieved by two types of observations: (1) $360^\circ$ point cloud input from a LiDAR (naturally continuous) or (2) $360^\circ$ cylindrical camera.
However, (1) may not seamlessly incorporate semantic information from RGB images, and (2) is hard to obtain and process.
Thus, we need to relax this requirement of $\sotwo$-transformable input modality. As a solution, we introduce a learnable layer $\texttt{lift}$ that can map camera images from different views to a $\sotwo$-transformable feature. This enhances our ability to exploit symmetry in the planning process. %

\begin{figure}[!t]
    \vspace*{0.15cm}
    \centering
    \includegraphics[width=0.65\columnwidth]{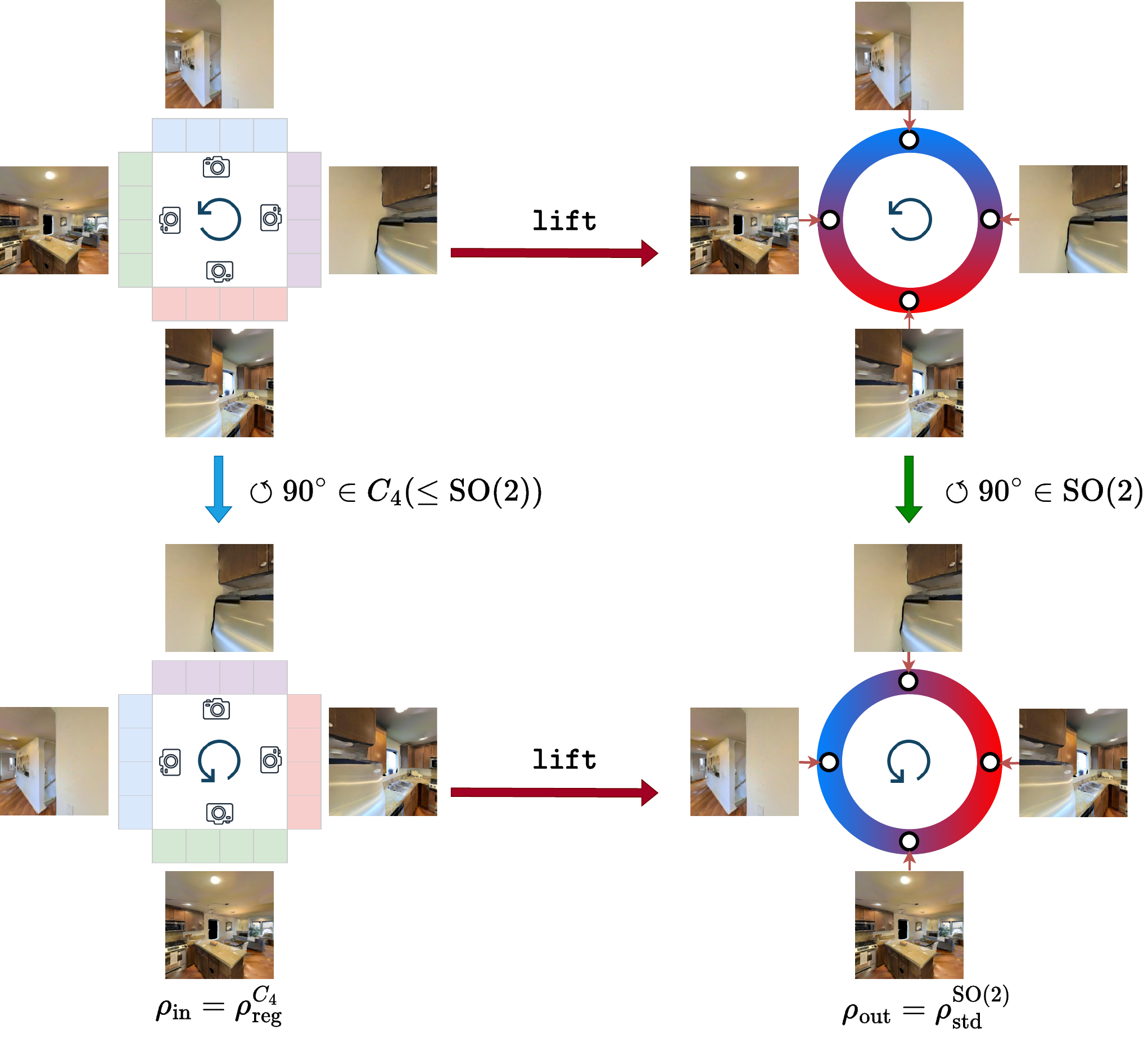}
  \caption{
  \small
 Our proposed $\texttt{lift}$ layer and its equivariance.
 }
  \vspace{-20pt}
    \label{fig: multi-cam}
\end{figure}

We visually illustrate this in Fig.~\ref{fig: multi-cam}.
For example, assume we have a robot equipped with four cameras facing north, east, west, and south (shown in the top left, as top down view).
The observation from this camera array could only be cyclically permuted by $\circlearrowleft 90^{\circ}$ (or reflected), shown in the bottom left using the blue arrow.
By using a equivariant learnable layer $\texttt{lift}$, it lifts the image features to become features on circle $S^1 \simeq \sotwo$ shown on the right.
They are transformable by $\sotwo$, as shown via green arrow.
We use small black circles to highlight that the feature at that point corresponds to that image.

Although the output is $\sotwo$-transformable, the left side is only $C_4$-transformable, so the layer $\texttt{lift}$ can only be \textit{restricted} to be $C_4$-equivariant. The restriction from $G=\sotwo$ to subgroup $H=C_4$ is called \textit{restricted representation.}
This layer is a special kind of equivariant induction layer \citep{howell2023equivariant}. It can lift features on a subgroup $H \leq G$ to a group $G$ and is $H$-equivariant.
Intuitively, it needs to satisfy the equivariance constraint only for $\circlearrowleft 90^{\circ} \in C_4$, which is a subgroup $C_4 \leq \sotwo$:
\begin{equation}
\begin{aligned}
    \texttt{lift}( \circlearrowleft 90^{\circ} \cdot \texttt{images} )
= \circlearrowleft 90^{\circ} \cdot \texttt{features},
\end{aligned}
\end{equation}
where we assume 4 images and output $\sotwo$ features, while it can be any group such that $C_4$ is its subgroup.

\footnotetext{
One solution for $D_4$ group is to use \textit{quotient} representations, but it not generally applicable for higher-degree rotations such as $D_8$ or infinitesimal rotations $\sotwo$.
}

\section{Experiments}

\begin{figure*}[!t]
    \vspace*{0.15cm}
    \centering
    \includegraphics[width=1.0\textwidth]{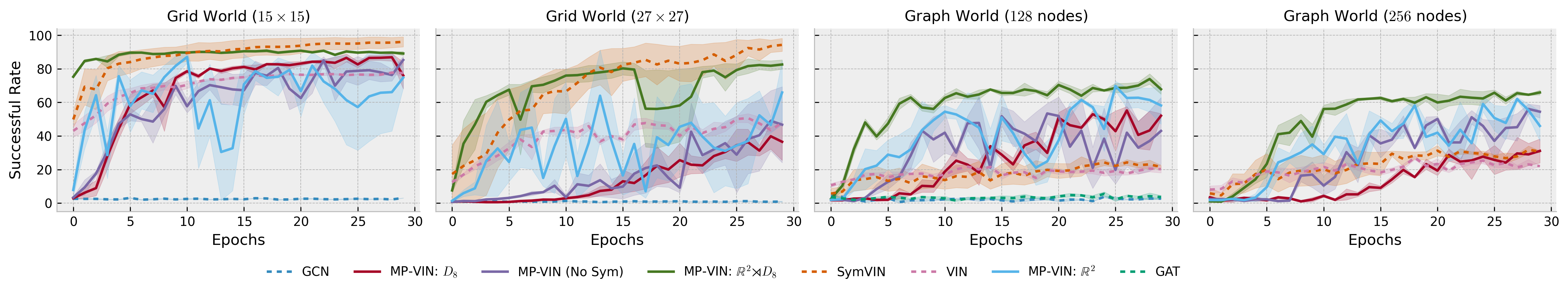}
    \vspace{-15pt}
    \caption{
    \small
    Learning curves on the Grid World experiments (left two) and the Graph World experiments (right two). The shadow area shows the standard error.
    Dashed lines are for non-MP-VIN methods (VIN, SymVIN, GCN-VIN, and GAT-VIN). 
     }
    \label{fig:grid-graph-result}
\end{figure*}

We evaluate our proposed approach MP-VIN and baselines on four different tasks. 
Among these tasks, we perform point goal navigation under different environments: 
known structured environments (\texttt{Grid World}), known unstructured environments (\texttt{Graph World}), unknown structured environments (\texttt{Miniworld}), and unknown unstructured environments (\texttt{Miniworld-Graph}).

\textbf{Methods.}
We experiment four variants of our methods, with or without translation ($\sR^2$) or rotation/reflection (using $G=D_8 \leq \otwo$) equivariance: No-Sym, $D_8$, $\sR^2$, and $\sR^2 \rtimes D_8$.
We use two grid-based methods: VIN \cite{tamar_value_2016} and SymVIN \cite{zhao_integrating_2022} (with $D_4$-equivariance).
These methods are grid-based; therefore, we apply several modifications, including pre-processing and post-processing, to ensure fair comparisons. These modifications are detailed in the later sections.
We also replace and compare our message passing module with the Graph Convolutional Networks \cite{kipf_semi-supervised_2017} (GCN-VIN) and Graph Attention Networks \cite{velickovic_graph_2018_hongyu} (GAT-VIN).

\subsection{Planning on known maps: Grid World}
\label{sec: grid-world}
\textbf{Setting.}
In this task, we randomly generate synthetic mazes with size $m \times m$ (\texttt{Grid World}). We validate the performance on two different sizes $m \in \{15,27\}$. Each cell on the maze map is represented as occupied (0) or unoccupied (1). There are four actions available for each cell on the map: north, east, west, and south. We randomly select a goal on the map and generate a goal map, where the cell containing the goal is marked as 1. Each cell is labeled by the ground-truth action using Dijkstra's algorithm.

To apply our planners for graphs, we transform the grid representation into connectivity graphs \cite{niu_generalized_2017_hongyu}.
Each \textit{node} of the graph is a cell in the 2D grid and is associated with a 4-D \textit{node feature vector}, which has (1) $(x,y)$ location, (2) whether the node is an obstacle, and (3) whether the node is the goal.
Any two nodes are connected by an \textit{edge} if they are neighbors on the grid map, i.e., obstacles are not connected.

\textbf{Results.}
MP-VIN with $\sR^2 \rtimes D_8$ symmetry demonstrates faster learning efficiency than its graph-based variants and VIN (the left of Fig.~\ref{fig:grid-graph-result}). We surprisingly find that MP-VIN with $\sR^2 \rtimes D_8$ has much smoother learning curves than its variants without $D_8$ symmetry ($\sR^2$ and No-Sym). This indicates that injecting Euclidean symmetry may improve the loss landscape. In terms of absolute performance gain, by adding $D_8$ symmetry to MP-VIN with $\sR^2$, we obtain another 0.69\% and 12.07\% success rate on the $15 \times 15$ and $27 \times 27$ mazes, respectively.
However, it is still outperformed by SymVIN, which uses steerable 2D convolution to process the input.
It is reasonable as it directly uses the regular grid structure, while our graph version can handle unstructured grpahs and is more expressive, while we apply both of them on grid maps.
When the map size increases, MP-VIN with $\sR^2 \rtimes D_8$ symmetry demonstrates the second-least performance degradation, showing better generalization to larger maps.

\subsection{Planning on known graphs: Graph World}
\label{sec: graph-world}
\textbf{Setting.}
We validate the performance of our approach in unstructured graph environments (\texttt{Graph World}). We follow the setup of \cite{niu_generalized_2017_hongyu} to generate the random graphs. We randomly generate $N$ nodes, each with coordinates between $(0,0)$ and $(m,m)$. These nodes are connected using a KNN graph. We randomly select some nodes as the obstacle nodes, and one node as the goal node.

To verify the performance of grid-based approaches in Graph World, we discretize the environment~\cite{niu_generalized_2017_hongyu}. We round down the coordinates of each node to map it to a cell on the grid. The obstacle feature is carried over to the grid. Ultimately, we verify its performance on the graph by transforming the discrete actions into continuous actions (represented in $[x,y]$ coordinate). For example, \texttt{north} is transformed into $[1,0]$, and \texttt{east} is transformed into $[0,1]$. 

\textbf{Results.}
MP-VIN with $\sR^2 \rtimes D_8$ symmetry demonstrates the strongest performance in this task in terms of learning efficiency and smoothness of learning curve (Fig.~\ref{fig:grid-graph-result}).
This is because the graphs generally do not have regular structure, i.e. four neighbors only in four directions.
Thus, all methods encounter performance degradation, while grid-based methods struggle more in such unstructure graphs.

\begin{figure}[h]
  \begin{center}
    \includegraphics[width=\columnwidth]{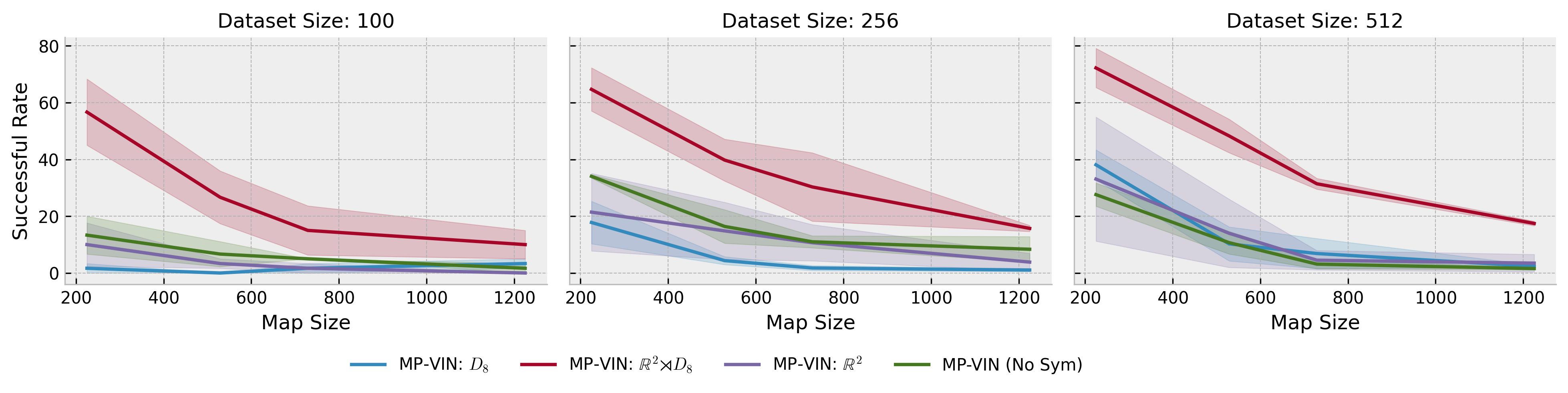}
  \end{center}
  \vspace{-1em}
  \caption{
  \small
  \textbf{Data efficiency and size generalization}. We demonstrate data efficiency across 100, 256, and 512 training samples. For models trained on each dataset, we show size generalization by training them on the smallest size and directly testing them on larger ones.}
  \vspace{-5pt}
    \label{fig: generalization}
\end{figure}
\textbf{Data Efficiency.}
As shown in Fig. \ref{fig: generalization}, we evaluate the data efficiency by assessing the model tranined on varied dataset sizes (100, 256, 512 samples).
Even when trained with only 100 samples, our approach, incorporating E(2) symmetry, consistently outperforms the baselines (w/o E(2) symmetry) trained with 512 samples. 
These results confirm the substantial gains in data efficiency achieved by leveraging Euclidean symmetry.

\textbf{Size Generalization.}
Fig. \ref{fig: generalization} illustrates our evaluation of size generalization ability.
We train our model on a small graph consisting of 225 nodes and subsequently test it on larger graphs without any further fine-tuning.
Remarkably, even as the complexity of the environment increases, our approach consistently outperforms the compared methods by a large margin.
This underscores the added value of incorporating E(2) symmetry in enhancing the model's generalizability to diverse environments.

\begin{table*}[t]
    \vspace*{0.15cm}
 \centering
     \caption{
     \small
Averaged test success rate (\%) with standard deviation. The best result is \textbf{bolded}. The second-best result is \underline{underlined}.
     }
    \label{tab:all-table} 
    \small
    \begin{tabular}{l|ll|ll|ll}
    \toprule

    \multirow{2}{*}{Method} & \multicolumn{2}{c}{Grid World} & \multicolumn{2}{c}{Graph World} & \multicolumn{2}{c}{Miniworld} \\  
                          & $15 \times 15$ & $27 \times 27$ & $128$ nodes & $256$ nodes & Grid & Graph \\

    \midrule
 VIN \citep{tamar_value_2016}  & $ 78.51 {\color{gray}\scriptstyle \pm 1.81} $ & $ 50.15 {\color{gray}\scriptstyle \pm 3.94} $ & $ 18.75 {\color{gray}\scriptstyle \pm 1.95} $ & $ 20.09 {\color{gray}\scriptstyle \pm 6.87} $ & $ 57.14 {\color{gray}\scriptstyle \pm 8.92} $ & $ 18.90 {\color{gray}\scriptstyle \pm 2.87} $ \\
 SymVIN \citep{zhao_integrating_2022} & $ \textbf{95.85} {\color{gray}\scriptstyle \pm 5.02} $ & $ \textbf{93.73} {\color{gray}\scriptstyle \pm 7.33} $ & $ 24.40 {\color{gray}\scriptstyle \pm 2.11} $ & $ 27.53 {\color{gray}\scriptstyle \pm 4.73} $ & $ \textbf{91.67} {\color{gray}\scriptstyle \pm 2.58} $ & $ 27.98 {\color{gray}\scriptstyle \pm 4.34} $ \\

\midrule
MP-VIN (No Sym)  & $ 87.07 {\color{gray}\scriptstyle \pm 4.53} $ & $ 55.99 {\color{gray}\scriptstyle \pm 39.56} $ & $ 63.10 {\color{gray}\scriptstyle \pm 17.26} $ & $ 54.76 {\color{gray}\scriptstyle \pm 3.29} $ & $ - $ & $ - $ \\
MP-VIN: $D_8$  & $ 87.19 {\color{gray}\scriptstyle \pm 2.78} $ & $ 38.53 {\color{gray}\scriptstyle \pm 16.17} $ & $ 52.38 {\color{gray}\scriptstyle \pm 5.02} $ & $ 32.89 {\color{gray}\scriptstyle \pm 3.47} $ & $ - $ & $ - $ \\
MP-VIN: $\sR^2$  & $ 90.81 {\color{gray}\scriptstyle \pm 1.02} $ & $ 72.45 {\color{gray}\scriptstyle \pm 31.94} $ & $ \underline{70.24} {\color{gray}\scriptstyle \pm 2.69} $ & $ \underline{58.33} {\color{gray}\scriptstyle \pm 5.93} $ & $ 79.76 {\color{gray}\scriptstyle \pm 24.06} $ & $ \underline{96.58} {\color{gray}\scriptstyle \pm 2.46} $ \\
MP-VIN: $\sR^2 \rtimes D_8$  & $ \underline{91.50} {\color{gray}\scriptstyle \pm 1.04} $ & $ \underline{84.52} {\color{gray}\scriptstyle \pm 6.04} $ & $ \textbf{72.17} {\color{gray}\scriptstyle \pm 5.08} $ & $ \textbf{61.90} {\color{gray}\scriptstyle \pm 5.33} $ & $ \underline{90.89} {\color{gray}\scriptstyle \pm 1.63} $ & $ \textbf{96.96} {\color{gray}\scriptstyle \pm 1.00} $ \\
        
        \bottomrule
    \end{tabular}
    \vspace{-10pt}
\end{table*}

\begin{figure}[h]
  \begin{center}
    \includegraphics[width=\columnwidth]{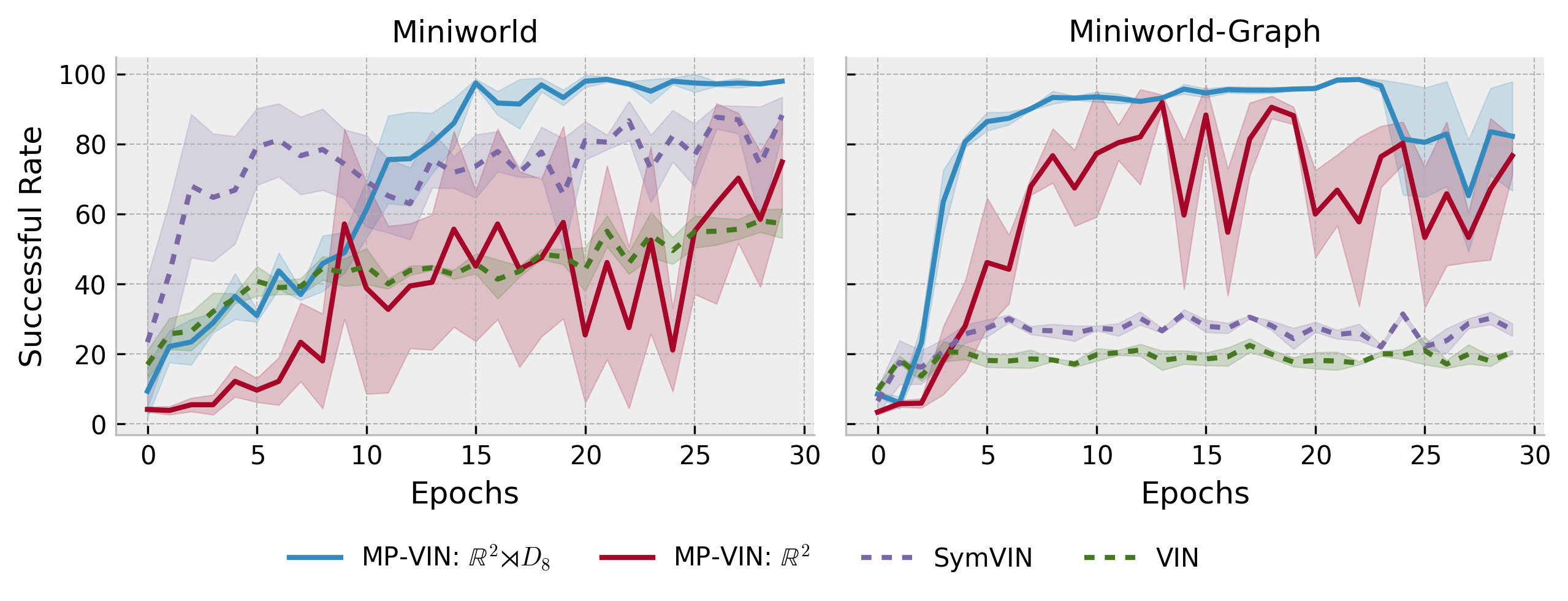}
  \end{center}
  \vspace{-1em}
  \caption{
  \small
  Learning curves on the \texttt{Miniworld} experiment (top) and \texttt{Miniworld-Graph} experiment.}
  \vspace{-15pt}
    \label{fig: miniworld}
\end{figure}

\subsection{Mapping and planning under unknown maps: Miniworld}

\label{sec: mini-world}
\textbf{Setting.}
We compare different methods in a more challenging visual environment (\texttt{Miniworld}), where the models learn mapping and planning simultaneously. We leverage the Miniworld simulator \cite{MinigridMiniworld23} to render the randomly generated maze into a 3D visual environment. Different from the Grid World, we are not given a map in this experiment. Instead, we use egocentric RGB observations. For each cell in the maze environment, we obtain the RGB images from the cameras facing four orientations (0$^{\circ}$, 90$^{\circ}$, 180$^{\circ}$, 270$^{\circ}$). We transform the dataset into a graph using the same approach as mentioned in Sec.~\ref{sec: grid-world}.
In order to estimate the map of the environment, we encode the visual observations into occupancy features using a mapper network \cite{lee_gated_2018_hongyu, zhao_integrating_2022}. 

\textbf{Results.}
Since this task is based on a $15 \times 15$ grid using visual observations, the experiment results are similar to the Grid World. MP-VIN with $\sR^2 \rtimes D_8$ symmetry demonstrates higher learning efficiency than MP-VIN with only $\sR^2$ symmetry and VIN. However, every method faces a performance drop due to the mapping uncertainty. We observe that the performance gap of MP-VIN with $\sR^2 \rtimes D_8$ symmetry (0.61\%) is lower than that of SymVIN (4.18\%). Therefore, the performance gap between MP-VIN with $\sR^2 \rtimes D_8$ symmetry and SymVIN becomes narrower.

\subsection{Mapping and planning under unknown graphs: Miniworld-Graph}
\label{sec: mini-world-graph}

While Miniworld is a 3D-rendered visual environment, the state and action spaces are still discrete (grid-based). To show real-world feasibility, we aim at a more realistic setting in this experiment.

\textbf{Setting.}
We sample random navigation graphs ($256$ nodes) in the Miniworld environment. The edges between nodes represent navigability. Like the Miniworld experiment in Sec.~\ref{sec: mini-world}, each node contains a panoramic egocentric RGB observation facing four directions. We use a similar mapper to estimate the map from the visual observations. Differently, we estimate the occupancy graph instead of the occupancy grid. 

\textbf{Results.}
Similar to the Miniworld experiment on grid representation, we observe the MP-VIN with $\sR^2 \rtimes D_8$ symmetry has higher learning efficiency than MP-VIN with only $\sR^2$ symmetry. Grid-based approaches suffer in this task since it is hard for CNN to process the expressive unstructured environment, also indicated in the previous \texttt{Graph World} experiment.
In both Miniworld experiments, we observe that MP-VIN with $\sR^2 \rtimes D_8$ symmetry has a much smoother learning curve and lower variance than MP-VIN with only $\sR^2$ symmetry. This indicates that by adding $\sR^2$ symmetry, we could further optimize the network with better stability.

\subsection{Planning with semantic goal}
\label{sec: habitat}

\begin{figure}[!ht]
  \begin{center}
    \includegraphics[width=\columnwidth]{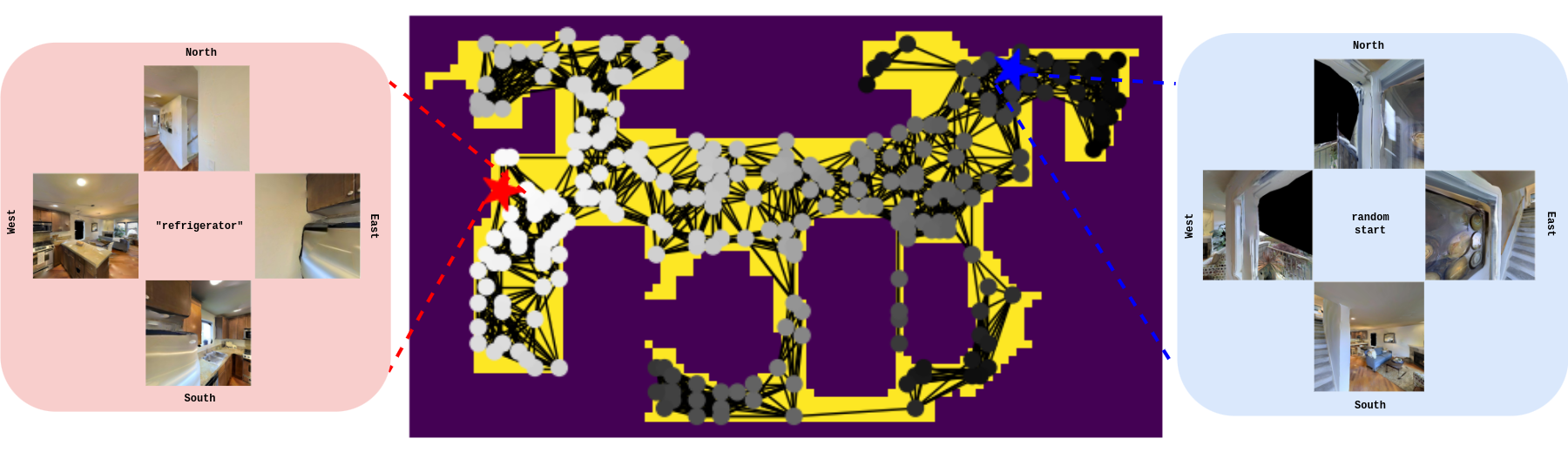}
  \end{center}
  \caption{\small \textbf{Visualization of our navigation environment.} On the left, we show the constructed geometric graph in an HM3DSem scene. The density of color represents the distance to the goal. On the right, we demonstrate the observation example on each node, which consists of four egocentric RGB images facing four directions.}
\end{figure}

\textbf{Setting.}
To confirm the validity of our approach in a more realistic setting, we perform a proof-of-concept semantic navigation task using the real-world collected Habitat-Matterport 3D semantics dataset (HM3DSem) \cite{yadav2023habitat}. Our model learns to seek an object in the environment given only RGB observations. In our experiment, we consider the most common object among all environments (``refrigerator''). We assume access to fully-observable environment information, \emph{i.e.} camera observations at any location.

 To achieve this, we randomly sample nodes in the navigatable areas using the Habitat simulator \cite{savva_habitat_2019_hongyu}. We construct a graph on the sampled nodes using a radius graph. The edges that lead to infeasible motion (\emph{e.g.} crossing the wall) are removed. We use the provided ground-truth object location to label the ground-truth action for each node. Note that the object location is unknown during testing. 
 We obtain four egocentric RGB images for each node, which are the only observations given to our model. 
 
For each scene, we randomly sample 20 graphs,
in which each graph contains 128 nodes. Each RGB image has the size of $3 \times H \times W$, and we set $H=W=128$ in our following experiments. We extract image features ($d=128$) from the RGB observation using ResNet-34. During training, we freeze the entire ResNet except for the last output layer. The image features are fed into the differentiable planner to generate the optimal plan.

\textbf{Results.}
We utilize MP-VIN without symmetry as the baseline, and compare MP-VIN with $\sR^2 \rtimes C_4$ symmetry. The result is shown in Tab. \ref{tab: habitat}. Our findings demonstrate that incorporating $C_4$ symmetry into MP-VIN leads to an improvement of 4.57\% in the success rate.

\begin{table}[!h]
 \centering
 \vspace{-5pt}
     \caption{
Performance in semantic navigation task.
     }
    \label{tab: habitat} 
    \small
    \begin{tabular}{l|ll}
    \toprule
    Method & Successful Rate (\%) \\
    \midrule
 MP-VIN (No Sym)  & $ 69.70 {\color{gray}\scriptstyle \pm 1.07} $  \\  
 MP-VIN: $\sR^2 \rtimes C_4$  & $ \textbf{74.27} {\color{gray}\scriptstyle \pm 3.12} $ \\       
        \bottomrule
    \end{tabular}
    \vspace{-15pt}
\end{table}

\section{Conclusion and Discussion}
In this letter, we explored the applicability of exploiting Euclidean symmetry within the context of a navigation planner. We contributed a novel equivariant differentiable planner. The effectiveness of the proposed approach is extensively assessed across four distinct tasks involving structured and unstructured environments, with known and unknown maps. The empirical findings demonstrate a significant enhancement in learning efficiency when Euclidean symmetry is integrated into 2D navigation planning. Furthermore, the results indicate that leveraging Euclidean symmetry yields more stable optimization and yields superior overall performance. In the future, we hope to extend our work to navigation task that has higher dimension, such as semantic navigation \cite{liang_sscnav_2021, chaplot_object_2020,chang_semantic_2020}.

\textbf{Limitations.}
Inheriting from VIN, our message passing planner also considers only fully-observable states and takes observations of all states as input at once \cite{tamar_value_2016,lee_gated_2018_hongyu,zhao_integrating_2022}, which is impractical in real-world navigation.
One potential direction is to consider partial observation, as done in \citep{ishida_towards_2022}.
Another potential direction is to combine with differentiable filter to counter the uncertainties~\cite{karkus_differentiable_2019_hongyu}.
To facilitate deployment on a real robot, it may be helpful to consider augmenting our state space with an additional orientation dimension~\cite{ishida_towards_2022}.

In this letter, our primary emphasis lies within the Euclidean group E(2). 
Nevertheless, in future works, potential performance improvement may be achieved by broadening the scope of the group employed, \eg incorporating the scaling and general linear group.

{
\small
\bibliography{custom,hongyu-custom,references-linfeng,references-add, references-manual} %
}

\clearpage
\appendix

\listoftodos

\section{Outline}

We provide mathematical preliminaries, extended discussion on method and derivation, experimental details, and additional results.

\section{Mathematical Preliminaries}
\label{sec:math-add}

We briefly introduce some preliminaries and used notations. For a more complete account, please check \citep{zhao_integrating_2022} for symmetry in planning and \citep{cohen_group_2016_hongyu,cohen_steerable_2016_hongyu,weiler_general_2021_hongyu,lang_wigner-eckart_2020_hongyu} for equivariant convolution networks.

\subsection{Group Representations}
Let $V$ be a vector space over $\mathbb{C}$. A \emph{representation} $(\rho , V)$ of $G$ is a map $\rho : G \rightarrow \Hom[V,V]$ such that 
\begin{align*}
\forall g , g' \in G, \enspace \forall v\in V, \quad   \rho( g \cdot g' )v =  \rho( g ) \cdot \rho(  g' )v
\end{align*}
Concisely, a group representation is a embedding of a group into a set of matrices. The matrix embedding must obey the multiplication rule of the group. Over $\mathbb{R}$ and $\mathbb{C}$ all representations break down into irreducible representations.

\paragraph{Restriction Representation}
Let $H \subseteq G$. Let $(\rho , V)$ be a representation of $G$. The restriction representation of $(\rho , V)$ from $G$ to $H$ is denoted as $\Res_{H}^{G}[ (\rho , V) ]$. Intuitively, $\Res_{H}^{G}[ (\rho , V) ]$ can be viewed as $(\rho , V)$ evaluated on the subgroup $H$ of $G$. Specifically, 
\begin{align*}
\forall h\in H, \enspace \forall v\in V, \quad \Res_{H}^{G}[ \rho ](h) v = \rho(h) v .
\end{align*}

\subsection{Group Convolution and Steerable Convolution}

\todoilzlf{check and adopt}

The understanding of this point helps to understand how a group acts on various feature fields and the design of state space for path planning problems.
We use the discrete group $p4 = \mathbb{Z}^2 \rtimes C_4$ as example, which consists of $\mathbb{Z}^2$ translations and $90^\circ$ rotations.

The group convolution with filter $\psi $ and signal $x$ on grid (or $\mathbf{p} \in \mathbb{Z}^2$), which outputs signals (a function) on group $p4$
\todozlf{use discrete version for here, from SymPlan, can use cont ver; harder to understand}
\begin{equation}
	[\psi \star x](\mathbf{t}, r) := \sum_{\mathbf{p} \in \mathbb{Z}^2} \psi((\mathbf{t}, r)^{-1} \mathbf{p})\ x(\mathbf{p}).
\end{equation}

\todoilzlf{removed derivation for simplicity; move back some}

A group $G$ has a natural action on the functions over its elements; if $x: G \to \mathbb{R}$ and $g \in G$, the function $g.x$ is defined as $[g.x](h) := x(g^{-1} \cdot h)$.

For example: The group action of a rotation $r \in C_4$ on the space of functions over $p4$ is
\begin{equation}
	[r.y](\mathbf{p}, s) := y(r^{-1} (\mathbf{p}, s)) = y(r^{-1}\mathbf{p}, r^{-1}s),
\end{equation}
where $r^{-1}\mathbf{p}$ spatially rotates the pixels, $r^{-1}s$ cyclically permutes the 4 channels.

The $G$-space (functions over $p$4) with a natural action of $p4$ on it:
\begin{equation}
	[(\mathbf{t}, r).y](\mathbf{p}, s) := y((\mathbf{t}, r)^{-1} \cdot (\mathbf{p}, s)) = y(r^{-1}(\mathbf{p} - \mathbf{t}), r^{-1}s)
\end{equation}

The group convolution in discrete case is defined as 
\begin{equation}
	[\psi \star x](g) := \sum_{h \in H} \psi(g^{-1} \cdot h)\ x(h).
\end{equation}

The group convolution with filter $\psi $ and signal $x$ on $p4$ group is given by:
\begin{equation}
	  [\psi \star x](\mathbf{t}, r) := \sum_{s \in C_4} \sum_{\mathbf{p} \in \mathbb{Z}^2} \psi((\mathbf{t}, r)^{-1} (\mathbf{p}, s))\ x(\mathbf{p}, s).
\end{equation}

Using the fact
\begin{equation}
	\psi((\mathbf{t}, r)^{-1} (\mathbf{p}, s)) = \psi(r^{-1} (\mathbf{p} - \mathbf{t}, s)) = [r.\psi](\mathbf{p} - \mathbf{t}, s),
\end{equation}
the convolution can be equivalently written into
\begin{equation}
	[\psi \star x](\mathbf{t}, r) := \sum_{s \in C_4} \left( \sum_{\mathbf{p} \in \mathbb{Z}^2} [r.\psi](\mathbf{p} - \mathbf{t}, s)\ x(\mathbf{p}, s) \right).
\end{equation}

So $\left( \sum_{\mathbf{p} \in \mathbb{Z}^2} [r.\psi](\mathbf{p} - \mathbf{t}, s)\ x(\mathbf{p}, s) \right)$ can be implemented in usual shift-equivariant convolution \textsc{conv2d}.

The inner sum $\sum_{\mathbf{p} \in \mathbb{Z}^2}$ is equivalently for the sum in steerable convolution, and the outer sum $\sum_{s \in C_4}$ implement rotation-equivariant convolution that satisfies $H$-steerability kernel constraint.
Here, the outer sum is essentially using the \textit{regular} fiber representation of $C_4$.

In other words, group convolution on $p4 = \mathbb{Z}^2 \rtimes C_4$ group is equivalent to steerable convolution on base space $\mathbb{Z}^2$ with the fiber group of $C_4$ with regular representation.

\section{Extended Method and Derivation}
\label{sec:add-derivation}

\subsection{Details on Pipeline}
\label{sec: appendix-details-pipeline}

We provide an example code for MP-VIN in the listing \ref{lst:mp-vi} and for MP-VIN with $\sR^2 \rtimes D_8$ symmetry in listing \ref{lst:symmp-vi}. \texttt{r\_mp} and \texttt{q\_mp} are our message passing layer. \texttt{r\_equiv\_mp} and \texttt{q\_equiv\_mp} are their respective equivariance implementations. We refer readers to our code repository for the full implementation once the paper is accepted.

\begin{figure}
\noindent\begin{minipage}{.48\textwidth}
\begin{lstlisting}[language=Python, basicstyle=\ttfamily\tiny, label={lst:mp-vi}, caption=The main value iteration procedure for MP-VIN.]
# Input: graph, #iterations K




# the first two dimensions are spatial coordinates
# the last remaining dimensions are features
pos = graph[:, :2]
feat = graph[:, :2]





r = r_mp(pos, feat)

# Init value function V
v = torch.zeros(r.size())


for _ in range(K):
   # Concat and convolve V with P
   rv = torch.cat([r, v], dim=1)
   q = q_mp(pos, rv)
   
   # Max over action channel
   # N represents node number
   # > Q: (batch_size x N) x q_size
   # > V: (batch_size x N) x 1
   v, _ = torch.max(q, dim=1)

# Output: 'q' (to produce policy map)
\end{lstlisting}
\end{minipage}\hfill
\begin{minipage}{.48\textwidth}
\begin{lstlisting}[language=Python, basicstyle=\ttfamily\tiny, label={lst:symmp-vi}, caption=The equivariant value iteration procedure for MP-VIN with symmetries.]
# Input: graph, #iterations K

from e2cnn.nn import GeometricTensor
from e2cnn.nn import tensor_directsum

# the first two dimensions are spatial coordinates
# the last remaining dimensions are features
pos = graph[:, :2]
feat = graph[:, :2]
pos_geo = GeometricTensor(pos, type=field_type_pos)
feat_geo = GeometricTensor(feat, type=field_type_feat)

r_geo = r_equiv_mp(pos_geo, feat_geo)

# Init V and wrap V in e2cnn 'geometric tensor'
v_raw = torch.zeros(r_geo.size())
v_geo = GeometricTensor(v_raw, field_type_v)

for _ in range(K):
   # Concat (direct-sum) and convolve V with P
   rv_geo = tensor_directsum([r_geo, v_geo])
   q_geo = q_equiv_mp(pos_geo, rv_geo)

   # Max over group channel
   # N represents node number
   # > Q: (batch_size x N) x (|G| * q_size)
   # > V: (batch_size x N) x (|G| * 1)
   v_geo = q_max_pool(q_geo)
   
# Output: 'q_geo' (to produce policy map)
\end{lstlisting}
\end{minipage}
\end{figure}

\subsection{Equivariance of Message Passing for Value Iteration}
\label{sec: theorem_1}

\textbf{\textit{Theorem 1}}
\textit{The Bellman operator is equivariant under Euclidean group $\mathrm{E}(2)$.}

\textit{Proof.}
The proof is analogous to the proof for discrete case in \citep{zhao_integrating_2022} and equivariant convolution in \citep{weiler_general_2021_hongyu}.

For any group element $g \in \mathrm{E}(d) = \sR^d \rtimes \mathrm{O}(d)$, we transform the Bellman (optimality) operator step-by-step and show that it is equivariant under $\mathrm{E}(d)$:
\begin{align}
&L_g \left[ \gT [V] \right](\vs) \\
& \stackrel{\text{(1)}}{=} \gT [V](g^{-1}\vs) \\
& \stackrel{\text{(2)}}{=} \max_{\va} R(g^{-1}\vs, \va) + \int d\vs' \cdot P(\vs' \mid g^{-1} \vs, \va) V(\vs') \\
& \stackrel{\text{(3)}}{=} \max_{\bar \va} R(g^{-1}\vs, g^{-1} \bar \va) + \int d(g^{-1} \bar{\vs}) \cdot P(g^{-1} \bar{\vs} \mid g^{-1} \vs, g^{-1}\va) V(g^{-1}  \bar{\vs}) \\
& \stackrel{\text{(4)}}{=} \max_{\bar \va} R(\vs, \bar \va) + \int d(g^{-1} \bar{\vs}) \cdot P(\bar{\vs} \mid \vs, \va) V(g^{-1} \bar{\vs}) \\
& \stackrel{\text{(5)}}{=} \max_{\bar \va} R(\vs, \bar \va) + \int d\bar{\vs} \cdot P(\bar{\vs} \mid \vs, \va) V(g^{-1} \bar{\vs}) \\
& \stackrel{\text{(6)}}{=} \gT [ L_g [V]](\vs) 
\end{align}

For each step:
\begin{itemize}
\item (1) By definition of the (left) group action on the feature map $V: \gS \to \sR$, such that $g\cdot V(\vs) = \rho_0(g) V(g^{-1} \vs) = V(g^{-1} \vs)$. Because $V$ is a scalar feature map, the output transforms under trivial representation $\rho_0(g) = \mathrm{Id}$.
\item (2) Substitute in the definition of Bellman operator.
\item (3) Substitute $\va = g^{-1} (g \va) = g^{-1} \bar \va$. Also, substitute $g^{-1} \bar \vs = \vs'$.
\item (4) Use the symmetry properties of Geometric MDP: $P(\vs' \mid \vs, \va) = P(g \cdot \vs \mid g \cdot \vs, g \cdot \va)$ and $R (\vs, \va) = R(g \cdot \vs, g \cdot \va)$.
\item (5) Because $g \in \mathrm{E}(2)$ is isometric transformations (translations $\sR^2$, rotations and reflections $\mathrm{O}(2)$) and the state space carries group action, the measure $ds$ is a $G$-invariant measure $d(gs) = ds$. Thus, $d \bar{\vs} = d(g^{-1} \bar{\vs})$. 
\item (6) By the definition of the group action on $V$.
\end{itemize}

The proof requires the geometric graph of the navigation MDP to have Euclidean symmetry and the state space carries a group action of Euclidean group.
Therefore, the Bellman operator of a Geometric MDP is $\mathrm{E}(d)$-equivariant.

\subsection{Equivariant Lifting Layer}
\label{subsec:lifting-layer-add}

\todoil{check detail completeness}

Suppose the number of cameras/images for each position $\vx \in \sR^2$ in $M$ is $K$, so the multi-view image $I = (I_1,  \ldots, I_K)$ belongs to $\sR^{K \times H \times W}$.
For example, if $K=4$, we write it as $I = ( I_1, I_2, I_3, I_4)$.
For this particular case, rotating a robot can only be $90^\circ \cdot k$ rotations from $C_4$ (cyclic group with $\nicefrac{k \cdot 2 \pi}{4}$ rotations), which corresponds to cyclically permuting images in $I$ (by \textit{regular representation} \editcorl{$\rho_{\text{reg}}$}) of $C_4$: $\rho_{\text{reg}}(\circlearrowleft 90^\circ) \cdot I = \rho_{\text{reg}}(\circlearrowleft 90^\circ) \cdot ( I_1, I_2, I_3, I_4) = (I_4, I_1, I_2, I_3)$.

However, this blocks us from using higher-order symmetry $G$ in later planning network.
To this end, we propose using a \textit{trainable equivariant layer} to \textit{lift} the $H$-features to $G$-features, which is to \textit{induce} how multi-view images $I = (I_1,  \ldots, I_K)$ are transformed under a larger group\footnotemark{} $G \geq H$.
\todoilzlf{what is H-feature and G-feature - planner is $G$-equivariant, image encoder is $H$-equivariant, so we want entire network to be $G$-equivariant}
Such \textit{induction} layer needs to satisfy the steerable kernel constraint with input and output representation:
\begin{align}
\rho_\text{in} = \rho_{\text{reg}}^{C_K} &,
\quad
\rho_\text{out} = \mathrm{Res}_{C_K}^{G} \left[\rho_{\text{reg}}^{G} \right], \\
\texttt{lift}( \rho_{\text{reg}}^{C_K}(g) \cdot \texttt{images} )
&=
\mathrm{Res}_{C_K}^{G} \left[\rho_{\text{reg}}^{G} \right] (g) \cdot \texttt{features},
\end{align}
where $\rho_\text{in}$ is (\eg, regular) representation of $C_K$ and $\rho_\text{out}$ is restricted representation ($\mathrm{Res}^G_H$) of (\eg, regular) representation of $\sotwo$ to $C_K$.
The restricted representation limits the equivariant constraints to only the subgroup $H=C_K$.

\todonote{added a bit more background}
\todoilzlf{What is Res - I need to rewrite this, give more context}
Intuitively, this layer only enforces $H$-equivariant ($H =C_K$) but not $G$: cyclically permuting images $\circlearrowleft 90^{\circ}$ should result in $\circlearrowleft 90^{\circ}$ rotated output.
However, it outputs $G$-equivariant features, thus proceeding layers are able to apply $G$-equivariance.
In implementation, we use discrete subgroup such as $D_8 \leq \otwo$, since continuous groups do not have finite-dimensional regular representation to permute $I =  (I_1, \ldots)$.

\footnotetext{
This requires $C_K$ to be a subgroup of $G$, which is always the case for $G = \sotwo$ and for $G = C_{K'}$ or $D_{K'}$ when $K'$ is divisible by $K$.
}

\subsection{Implementation of MLPs}
\label{subsec:implement-mlp}
In this section, we introduce message passing operation, which is the basis of our MP-VIN and its variants.
In an undirected graph $G$, the node $u$ contains node feature $x_u$. The edge connecting nodes $(u,v)$ contains edge feature $e_{uv}$.
To pass the message along the edges for $T$ steps, $T$ message passing layers are stacked together.
\begin{equation}
    m_u^{t+1} = \bigoplus_{v \in N(v)} \texttt{propagate}_\theta(h_u^t, h_v^t, e_{uv}),
\end{equation}

here $m_u^{t+1}$ represents the message at node $u$ at time step $t+1$, and $h_u^t$ is the hidden state. $\bigoplus$ could be any permutation invariant and differentiable functions such as summation $\sum$. The message at each node is then updated by

\begin{equation}
    h_u^{t+1} = \texttt{update}_\theta(h_u^t, m_u^{t+1}) .
\end{equation}

In the previous equations, $\texttt{propagate}_\theta$ and $\texttt{update}_\theta$ are two differentiable functions (following the notations defined in our Methodology Section \ref{sec: methodology}), and MLPs are commonly used.

\todoil{briefly introduce how to implement those 2 MLPs}

\section{Experimental Details}
\label{sec: appendix-exp_details}

In the experiments, we optimize all approaches using RMSprop optimizer. This is the standard training setting in VIN and its variants. We also try the Adam optimizer, but the performance gain is not statistically significant. We choose the learning rate of 1e-3 and the batch size of 32.

For the dataset size, we choose 10K/2K/2K for the train/val/test dataset for the Grid World experiment, and 1K/200/200 for the remaining experiments.

\subsection{Planning on known maps}
In this experiment, we choose the value iterations $K=\{20,40\}$ for $m=\{15,27\}$ respectively.
The value is propagated in the network using either convolution (grid-based) or message passing layer (graph-based). Therefore, when the map size is larger, it takes more iterations to propagate.

\subsection{Planning on known graphs}
\label{sec: appendix-known-graph}
We set the iteration numbers $K=20$ on both $128$ nodes and $256$ nodes task. Among all nodes, we randomly select 10\% of the nodes as the obstacle nodes in our experiments.

To allow fair comparisons with grid-based approaches in Graph World, we adopt the following steps to discretize the environment \cite{niu_generalized_2017_hongyu}. We round down the coordinates of each node to the nearest integer, such that each node in the graph is mapped to a cell on the grid \footnote{Each node in the graph is guaranteed to find its corresponding cell location in a $m \times m$ map.}. Each obstacle node is marked as occupied on its respective cell location in the grid, and the remaining cells are marked as unoccupied. Since the rounding operation is a subjective mapping, we consider a cell to be occupied if any of the nodes are occupied. We train the grid-based approaches using this transformed grid, and verify its performance on the graph by transforming the discrete actions into continuous actions (represented in $[x,y]$ coordinate).

In our approach, the output of our network corresponds to the relative spatial coordinate $[\Delta x, \Delta y]$, enabling transitions within the navigation graphs. This differs from grid-based methods that employ NEWS actions ($\uparrow, \leftarrow, \downarrow, \rightarrow$). To facilitate a fair comparison with grid-based methods on geometric graphs, we discretize the graph into a grid space. Furthermore, to align with the output of grid-based methods, we also discretize the actions. Specifically, each action is matched with its nearest neighbor among $\{[1,0],[-1,0],[0,1],[0,-1]\}$. Subsequently, the resulting 2D vector is translated into the corresponding NEWS action.

\subsection{Mapping and planning under unknown maps}
\label{sec: appendix-miniworld}

\paragraph{Maze setting.}
In our experiment, we render the 3D visual environment using the Miniworld environment. In our experiment, we test on the map size of $15 \times 15$.

\paragraph{Grid mapper.}
We implement of baseline mapper network baseline on \citeauthor{lee_gated_2018_hongyu}. The mapper is a fully convolutional network and consists of two parts: an image encoder and a map decoder. The image encoder encodes the panoramic observation into a latent vector, and the map decoder predicts the map from it.

The image encoder takes $m \times m$ panoramic RGB observations facing four directions. Each RGB image has a resolution of $32\times32\times3$. Therefore, the shape of the input tensor is $m \times m \times 4 \times 32 \times 32 \times 3$. We pass the input tensor through three CNNs: $[(32,10,4,4), (64,5,2,2), (256,4,1,0)]$. Each tuple represents the filter number, kernel size, stride number, and padding. We transform the output into a $15 \times 15$ map with a feature dimension of $256$. 

The map decoder reconstructs the map using the feature on each cell. It consists of two CNNs: $[(32, 3, 1, 1), (1, 3, 1, 1)]$. We use Sigmoid as our last activation function to predict a map with binary value.

\subsection{Mapping and planning under unknown graphs}
\label{sec: appendix-miniworld-graph}
\paragraph{Graph mapper.}
The graph mapper shares a similar structure with the grid mapper but differs in the representation. The graph mapper consists of two components: an image encoder and a graph decoder.

The image encoder has the same structure as the one in the grid mapper. However, the input tensor to the image encoder has the shape of $(m \times m) \times 4 \times 32 \times 32 \times 3$.

The graph decoder reconstructs the occupancy graph from the latent vector. Therefore, we leverage two fully-connect layers with the size of $[32, 1]$. 

\section{Additional Results}

\subsection{Ablation: Symmetry Group}

\begin{figure}[!h]
    \vspace{-3em}
  \begin{center}
    \includegraphics[width=\columnwidth]{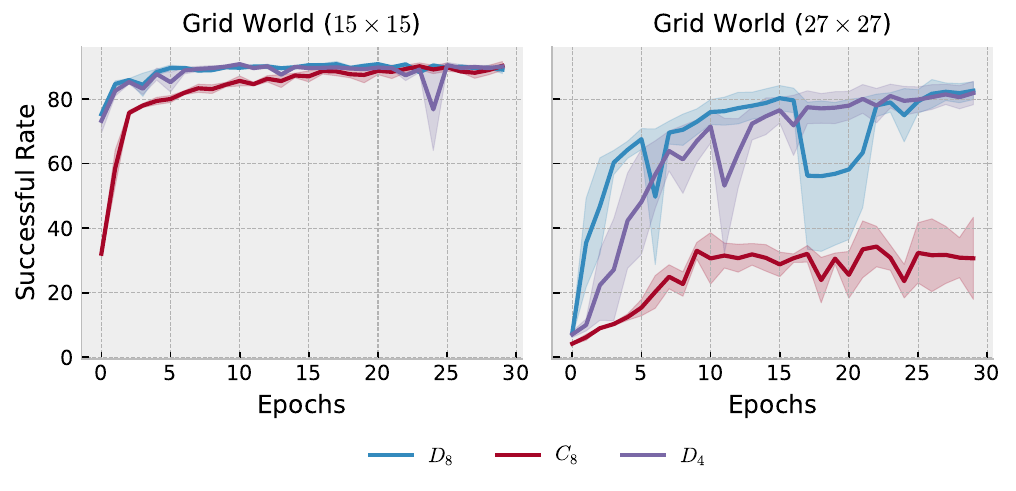}
  \end{center}
  \vspace{-1em}
  \caption{Learning curves on the \texttt{Miniworld} experiment (top) and \texttt{Miniworld-Graph} experiment.}
  \vspace{-1em}
    \label{fig: ablation-group}
\end{figure}

We perform this ablation study on the effectiveness of different choices of symmetry groups. Specifically, we study $C_8$, $D_4$, and $D_8$ group. We demonstrate the learning curves in Fig. \ref{fig: ablation-group} and Tab. \ref{tab:group-abla-table}. Our results show that $D_8$ group has an improvement over $D_4$ since it has more degrees of rotation symmetry. Even though $C_8$ and $D_4$ groups both have eight transformations, $D_4$ group performs significantly better in the $27 \times 27$ map. This is because the maze environment contains more reflection symmetry than rotation symmetry.

\begin{table}[!h]
 \centering
     \caption{
Averaged test success rate (\%) and standard deviation for our experiments.
     }
    \label{tab:group-abla-table} 
    \small
    \begin{tabular}{l|ll}
    \toprule
    \multirow{2}{*}{Group} & \multicolumn{2}{c}{Grid World} \\  
                          & $15 \times 15$ & $27 \times 27$ \\
    \midrule
 $C_8$  & $ 91.10 {\color{gray}\scriptstyle \pm 1.93} $ & $ 40.51 {\color{gray}\scriptstyle \pm 6.94} $ \\  
 $D_4$  & $ 90.59 {\color{gray}\scriptstyle \pm 1.00} $ & $ 82.11 {\color{gray}\scriptstyle \pm 6.29} $ \\       
 $D_8$  & $ \textbf{91.50} {\color{gray}\scriptstyle \pm 1.04} $ & $ \textbf{84.52} {\color{gray}\scriptstyle \pm 6.09} $ \\       
        \bottomrule
    \end{tabular}
    \vspace{-10pt}
\end{table}

\end{document}